%% file: acl_latex.tex
\pdfoutput=1

\documentclass[11pt]{article}

\usepackage[preprint]{acl}

\usepackage{times}
\usepackage{latexsym}

\usepackage[T1]{fontenc}

\usepackage[utf8]{inputenc}

\usepackage{microtype}

\usepackage{inconsolata}

\usepackage{graphicx}

\usepackage{hyperref}
\usepackage{url}            
\usepackage{booktabs}       
\usepackage{amsfonts}       
\usepackage{nicefrac}       
\usepackage{microtype}      
\usepackage{amsmath}
\usepackage{threeparttable}
\usepackage{graphicx}
\usepackage{multirow}
\usepackage{subfigure}
\usepackage{listings}
\usepackage{tikz}
\usepackage{colortbl}
\usepackage{xspace}
\usepackage{algorithm}
\usepackage{algpseudocode}
\usepackage{enumitem}
\usepackage{xcolor}         
\usepackage{tabularx} 
\usepackage{booktabs} 

\usepackage{amsthm}
\usepackage[english]{babel}
\theoremstyle{definition}
\newtheorem{definition}{Definition}
\newtheorem{remark}{Remark}

\lstdefinestyle{mystyle}{
    backgroundcolor=\color{backcolour},   
    commentstyle=\color{codegreen},
    keywordstyle=\color{magenta},
    numberstyle=\tiny\color{codegray},
    stringstyle=\color{codepurple},
    basicstyle=\ttfamily\footnotesize,
    breakatwhitespace=false,         
    breaklines=true,                 
    captionpos=b,                    
    keepspaces=true,                 
    numbers=left,                    
    numbersep=5pt,                  
    showspaces=false,                
    showstringspaces=false,
    showtabs=false,                  
    tabsize=2
}

\definecolor{codegreen}{rgb}{0,0.6,0}
\definecolor{codegray}{rgb}{0.5,0.5,0.5}
\definecolor{codepurple}{rgb}{0.58,0,0.82}
\definecolor{backcolour}{rgb}{0.95,0.95,0.92}
\definecolor{brickred}{rgb}{0.8, 0.25, 0.1}
\definecolor{midnightblue}{rgb}{0.1, 0.1, 0.44}
\definecolor{oceanboatblue}{rgb}{0.0, 0.47, 0.75}
\newcommand{\redbf}[1]{\bf{\textcolor{brickred}{#1}}}

\definecolor{lightgray1}{gray}{0.95}  

\newcommand{\circlednumber}[1]{%
    \tikz[baseline=(char.base)]{%
        \node[shape=circle,draw,inner sep=1pt] (char) {\scriptsize#1};%
    }%
}

\newcommand{\modelname}{{\sc TabGen-ICL}\xspace}
\title{\modelname: Residual-Aware In-Context Example Selection for Tabular Data Generation}

\author{
 \textbf{Liancheng Fang \textsuperscript{1}},
 \textbf{Aiwei Liu \textsuperscript{2}},
 \textbf{Hengrui Zhang \textsuperscript{1}},
 \textbf{Henry Peng Zou \textsuperscript{1}},
\\
 \textbf{Weizhi Zhang\textsuperscript{1}},
 \textbf{Philip S. Yu\textsuperscript{1}},
\\
 \textsuperscript{1}University of Illinois Chicago,
 \textsuperscript{2}Tsinghua University,
\\
\href{mailto:zhonghaoli@hkust-gz.edu.cn}{lfang87@uic.edu},
\href{mailto:xuminghu@hkust-gz.edu.cn}{liuaw20@mails.tsinghua.edu.cn},
\href{mailto:psyu@uic.edu}{psyu@uic.edu}
}

\begin{document}
\maketitle
\begin{abstract}
    Large Language models (LLMs) have achieved encouraging results in tabular data generation. However, existing approaches require fine-tuning, which is computationally expensive. This paper explores an alternative: prompting a fixed LLM with in-context examples. We observe that using randomly selected in-context examples hampers the LLM's performance, resulting in sub-optimal generation quality.
    To address this, we propose a novel in-context learning framework: \modelname, to enhance the in-context learning ability of LLMs for tabular data generation. 
    \modelname operates iteratively, retrieving a subset of real samples that represent the \textit{residual} between currently generated samples and true data distributions. This approach serves two purposes: locally, it provides more effective in-context learning examples for the LLM in each iteration; globally, it progressively narrows the gap between generated and real data.
    Extensive experiments on five real-world tabular datasets demonstrate that \modelname significantly outperforms the random selection strategy. Specifically, it reduces the error rate by a margin of $3.5\%-42.2\%$ on fidelity metrics. We demonstrate for the first time that prompting a fixed LLM can yield high-quality synthetic tabular data. 
    The code is provided in the \href{https://github.com/fangliancheng/TabGEN-ICL}{link}.
\end{abstract}

\input{Contents/1_introduction}

\input{Contents/2_related_works}

\input{Contents/3_method}

\input{Contents/4_experiment}

\input{Contents/5_conclusion}

\clearpage
\newpage
\input{Contents/6_limitations}

\clearpage 
\newpage 

\bibliography{references}

\newpage 
\appendix
\input{Contents/_appendix}

\end{document}

%% file: Contents/1_introduction.tex
\section{Introduction}

Tabular data, despite being one of the most prevalent data modalities in real-world applications \citep{benjelloun2020google}, often encounters several issues in practical use. These include imbalanced data categories \citep{cao2019learning}, privacy concerns \citep{gascon2016privacy} (as many tabular datasets contain sensitive personal information that cannot be directly shared), insufficient data quality \cite{lin2020missing}, and high data collection costs \citep{even2007economics}. Tabular generation is an important means to address these problems. Classic tabular generation methods such as GANs \citep{ctgan}, VAEs \citep{goggle}, and diffusion models \citep{stasy,codi,tabddpm,tabsyn} have two main limitations. First, they require large amounts of tabular data for training, which leads to a noticeable decline in performance in low-resource scenarios. This is particularly problematic considering that most real-world situations requiring tabular generation lack abundant data. Second, they need special preprocessing to handle heterogeneous data types, making them less flexible.

The rapid development of large language models (LLMs) brings new possibilities for solving table data generation problems with their powerful semantic understanding, reasoning, and generation capabilities. LLMs can understand and process various data types and structures without complicated data preprocessing, offering more flexible and principled solutions. Moreover, LLMs' few-shot learning ability may alleviate data scarcity issues, enabling excellent performance in low-resource scenarios. Previous works \citep{great, realtabformer, taptap, tabula, tabmt, Xu2024AreLN, Wang2024HARMONICHL} resort to fine-tuning general-purpose LLMs on target tables. While effective, fine-tuning requires substantial computational resources, making it inapplicable in resource-scarce scenarios. 

\begin{figure}[t!] 
    \centering

    \subfigure[No in-context Ex.]{\includegraphics[width=0.48\linewidth]{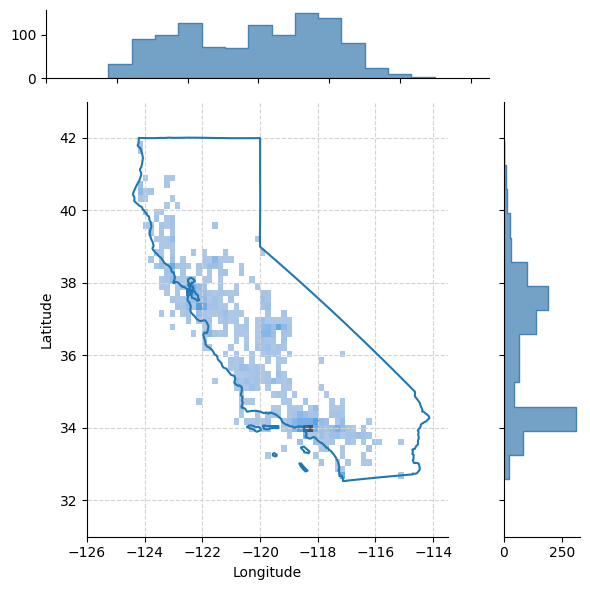}}
    \hfill
    \subfigure[Sampled in-context Ex.]{\includegraphics[width=0.48\linewidth]{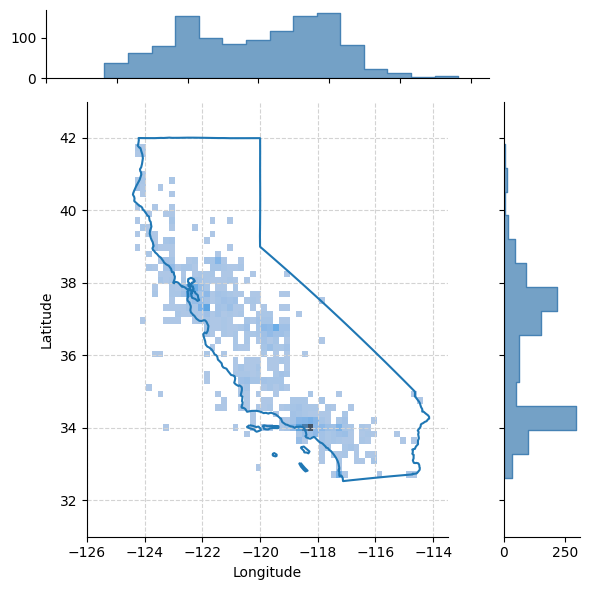}}
    
    \subfigure[Fixed in-context Ex.]{\includegraphics[width=0.48\linewidth]{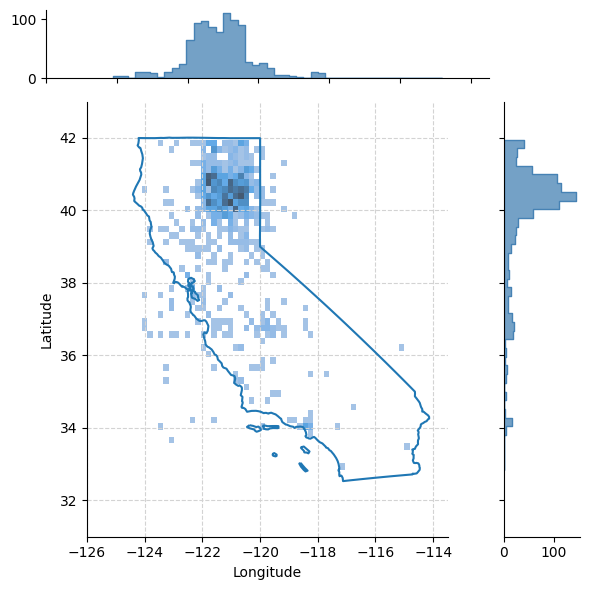}}
    \hfill
    \subfigure[Ground Truth]{\includegraphics[width=0.48\linewidth]{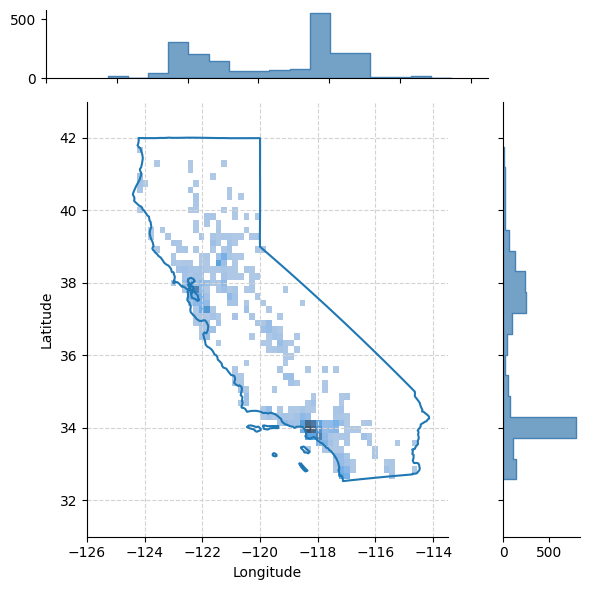}}
\caption{Comparison of samples generated with different in-context learning examples. Plots show the latitude and longitude coordinates of California housing, with the solid line representing the state boundary.
(a) 2000 samples generated by LLM with only the table header as input, \textbf{without} any in-context examples.
(b) 2000 samples generated by LLM, giving in-context examples sampled from the real dataset.
(c) 2000 synthetic samples generated by LLM, giving in-context examples with latitude and longitude in a fixed range. (d) 2000 samples from the ground truth training table.}
\label{fig:in-context-examples}
\vspace{-1em}
\end{figure}

In-context learning effectively solves such problems. By adding examples to the context, distribution characteristics can be provided to LLMs, guiding them to generate data that conforms to the target distribution without specific fine-tuning \citep{gao2023retrieval}. However, simple in-context learning strategies still face challenges. Figure \ref{fig:in-context-examples}(a) shows that even without in-context examples (see the full prompt at Appendix \ref{appendix:dummy_prompt}), LLMs can generate reasonable distributions, reflecting the influence of the LLM's pre-training distribution. Figure \ref{fig:in-context-examples}(b) demonstrates the strategy proposed by \citep{cllm}, which involves random sampling from Ground Truth as in-context examples. Although the generated results are closer to the Ground Truth shown in Figure \ref{fig:in-context-examples}(d) compared to Figure \ref{fig:in-context-examples}(a), they are still mainly influenced by the LLM's original distribution and struggle to fit the Ground Truth.

This phenomenon reveals the importance of choosing in-context examples. In this work, we propose \modelname, a dynamic in-context example selection method. 
Inspired by the observation in Figure \ref{fig:in-context-examples}(c), we found that using fixed range in-context examples leads to generated distributions closely mimicking those examples, significantly differing from the LLM's original distribution. This indicates in-context learning's ability to simulate distributions. By carefully selecting in-context examples, we can more effectively guide LLMs to generate distributions closer to the ground truth.
 
Central to our framework is the design of an automated strategy for selecting effective in-context examples while ensuring global consistency with the real data distribution. Our key idea is to utilize simple, discernible patterns in subsets of real samples, which can effectively guide LLMs in generating realistic tabular data. Specifically, \modelname identifies subsets of real samples that exhibit simple patterns and closely match the residual between the current generated data distribution and the real data distribution. This idea can be categorized as a novel residual-aware RAG technique, where we retrieve in-context examples based on the residual between the generated and real data distributions.

The residual-aware sampling measures the discrepancy between the generated and real data distributions, focusing on areas where the model needs improvement. This approach enables \modelname to progressively narrow the distribution gap while maintaining the use of easily learnable patterns in the in-context examples. Our sampling technique offers two key advantages: flexibility in selecting simple patterns for effective learning, and consistent generation through progressive distribution alignment. The contributions of this paper are as follows:
\begin{enumerate}[leftmargin=*]
    \item We propose \modelname, an in-context learning selection method that retrieves in-context examples by leveraging residual between currently generated samples and true data distributions.
    
    \item We conduct extensive experiments on five datasets, evaluated under three distinct groups of synthetic data evaluation metrics. Experiment results show that \modelname outperforms the previous in-context learning method by a margin of $3.5\%-42.2\%$ across multiple fidelity metrics. Notably, \modelname surpasses state-of-the-art deep generative models under the data-scarce scenarios.
\end{enumerate}

%% file: Contents/2_related_works.tex
\section{Related works} \label{sec:related_works}
\paragraph{Deep generative models for synthetic tabular data generation}
Generative models for tabular data have become increasingly important and have widespread applications~\cite {privacy, TabCSDI, privacy_health}. For example, CTGAN and TAVE~\citep{ctgan} deal with mixed-type tabular data generation using the basic GAN~\citep{gan} and VAE~\citep{vae} framework. GOGGLE~\citep{goggle} incorporates Graph Attention Networks in a VAE framework such that the correlation between different data columns can be explicitly learned. Recently, inspired by the success of Diffusion models in image generation, a lot of diffusion-based methods have been proposed, such as TabDDPM~\citep{tabddpm}, STaSy~\citep{stasy}, CoDi~\citep{codi}, and TabSyn~\citep{tabsyn}.

\paragraph{LLMs for synthetic data generation.}
Collecting high-quality training data for advanced deep-learning models is often costly and time-consuming. Researchers have recently explored using pre-trained large language models (LLMs) to generate synthetic datasets as a promising alternative. While LLMs have shown proficiency in generating high-quality synthetic text data, their ability to accurately replicate input data distributions at scale remains uncertain \citep{xu2024llms}. Studies like Curated-LLM \citep{cllm} demonstrate LLMs' effectiveness in augmenting tabular data in low-data scenarios, but their application to large-scale input data is still unclear. GReaT \citep{great}, another approach using GPT-2, generates synthetic tabular data but requires fine-tuning for each new dataset.

%% file: Contents/3_method.tex
\section{Preliminaries}
\label{preliminaries}
\paragraph{Notation.}  Tabular dataset refers to data organized in a tabular format with $N$ row and $D$ columns, where each row denotes a data record or sample, and each column denotes an attribute or feature. Each attribute can be either discrete (e.g. categorical) or continuous (e.g. real number $\mathbb{R}$). We use $\mathbb{P}(\boldsymbol{x})$ to denote the probability distribution of $\boldsymbol{x}$.

\paragraph{Data Setup.} We have access to a training dataset of $N$ samples: $\mathcal{D}_{train} = \{\boldsymbol{x}_i\}_{i=1}^N$, each sample $\boldsymbol{x}_i$ is \textit{i.i.d.} drawn from an unknown distribution $\mathbb{P}(\boldsymbol{x})$. 

\paragraph{Objective.} The goal is to generate a \textbf{new} dataset $\mathcal{D}_{syn}=\{\hat{\boldsymbol{x}}_{i}\}_{i=1}^N$ such that $\hat{\boldsymbol{x}}_i$ is \textit{i.i.d.} sampled from $\mathbb{P}(\boldsymbol{x})$. Direct copy of training data is not allowed.

\paragraph{Serialization.}
As LLMs primarily process text input, it is necessary to convert tabular data into a suitable textual format. There are many serialization formats for tabular data, such as JSON \citep{singha2023tabular}, Markdown \citep{sui2024table}, Sentences \citep{great}, etc. Notably, the JSON format is widely supported by LLMs, with models like GPT-4o capable of generating structured outputs in JSON format through constrained decoding \citep{liu2024we}.
Therefore, in this study, we adopt a JSON format to serialize tabular data. For instance, a row from a table containing three columns—\texttt{name} (categorical), \texttt{age} (numerical), and \texttt{city} (categorical)—is transformed into a JSON object: $\{\texttt{{name:`Alice', age:25, city:`New York'}}\}$. For a table comprising $N$ rows, the serialized data becomes a list of $N$ JSON objects. See Appendix \ref{lst:json} for the implementation of the JSON schema.
During each prompting iteration, \modelname retrieves a subset of these JSON objects to serve as in-context examples. This process will be elaborated upon in subsequent sections. 

\begin{figure*}[t!]
  \centering
  \includegraphics[width=\textwidth]{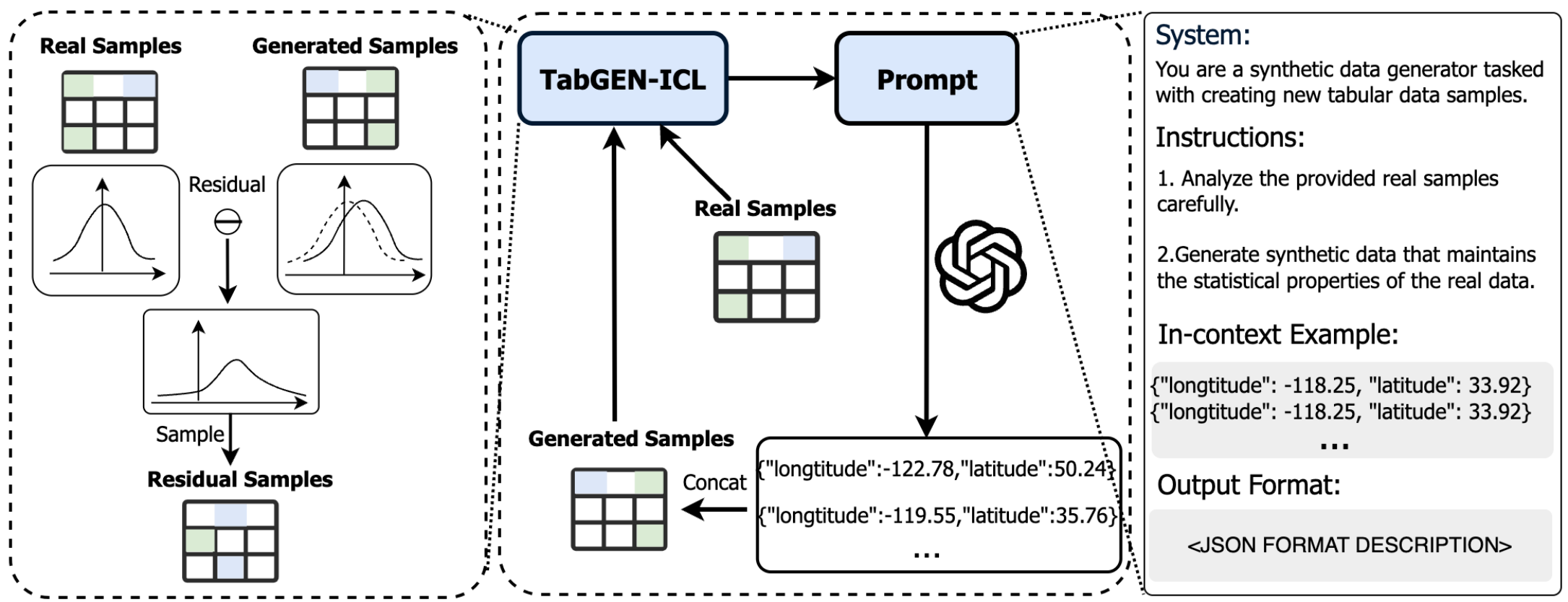}
  \caption{Overview of \modelname framework. We generate synthetic samples in batches, at each prompt iteration, \modelname retrieves a subset of real samples that acts as a \textit{residual} between the currently generated samples and the real data. The residual samples will be used as in-context examples to prompt LLMs in the next iteration. The full prompt template is available in the Appendix \ref{lst:prompt}.}
  \label{fig:overview}
\end{figure*}

\section{\modelname}
\label{method}
This section presents \modelname framework for tabular data generation. \modelname retrieves a subset of samples from the training dataset that satisfy two properties: 1) \textbf{Local}: at each prompting iteration, the LLMs can effectively extract patterns from the in-context examples; 2) \textbf{Global}: after enough iterations, the overall generated samples mimic the distribution of the real samples. In the following, we will introduce each component of \modelname in detail.

\subsection{LLM Generation with In-context Examples}
Our key observation is LLMs have strong prior distribution, and LLMs tend to generate samples following their prior distribution, neglecting the in-context examples, see Figure \ref{fig:in-context-examples}. Formally, given in-context examples, we assume the LLMs generate samples following a mixture distribution:
\begin{definition}[\textbf{LLM Generation Distribution}] \label{def:gen_dist}
Given the empirical distribution of in-context example: $\mathbb{P}_{ic}$. We define the LLMs generation distribution to be the following mixture of distributions:
\begin{equation} \label{eq:gen_dist}
\mathbb{P}_{gen} := \lambda \mathbb{P}_{llm} + (1-\lambda) \mathbb{P}_{ic}
\end{equation}
where $\mathbb{P}_{llm}$ is the prior distribution of LLMs, 
$\lambda\in[0,1]$. To sample from $\mathbb{P}_{gen}$, we first sample an index $z$ from a categorical distribution over $\{0,1\}$ with parameter $\lambda$, then sample from the corresponding distribution: 
\begin{equation} \nonumber
  \mathbb{P}_{gen}(\boldsymbol{x}) = \begin{cases} 
    \mathbb{P}_{llm}(\boldsymbol{x}) & \text{if } z = 1 \\
    \mathbb{P}_{ic}(\boldsymbol{x}) & \text{if } z = 0 
  \end{cases}
\end{equation} 
\end{definition}

Definition \ref{def:gen_dist} quantifies how the in-context examples steer the LLMs' generation from its own prior distribution towards the target distribution. Intuitively, the more in-context examples being provided, $\lambda$ will be closer to $0$, meaning the LLMs is more likely to generate samples following the empirical distribution of in-context examples. In practice, due to the limited context window of LLMs, only a small number of in-context examples can be provided, thus we expect $\lambda$ to be close to $1$. 

\subsection{In-context Examples Selection}
Recall our goal is to let LLMs generate samples that follow the same distribution as the training table, i.e. $\mathbb{P}_{gen} \approx \mathbb{P}_{train}$. It is tempting to choose the in-context examples by sampling from the empirical distribution of the training table, i.e. $\mathbb{P}_{ic} = \mathbb{P}_{train}$ \citep{cllm}. However, as the LLMs' generation is affected by the prior distribution $\mathbb{P}_{llm}$, the actual output distribution of LLMs would be $\mathbb{P}_{gen} = \lambda \mathbb{P}_{llm} + (1-\lambda) \mathbb{P}_{train}$, which is not our target distribution $\mathbb{P}_{train}$. Instead, a more plausible way is to select in-context examples s.t., when combined with $\lambda$ proportion of data generated from $\mathbb{P}_{llm}$, the resulting distribution is close to $\mathbb{P}_{train}$. In other words, the in-context examples can be understood as the \textit{residual} of $\mathbb{P}_{train}$ w.r.t. $\mathbb{P}_{llm}$. Formally, we introduce the definition of residual as follows:
\begin{definition}[\textbf{Residual}]
Let $\boldsymbol{X}$ be a set of $N$ $i.i.d.$ samples from a data distribution $\mathbb{P}(\boldsymbol{x})$, and let $\boldsymbol{Y}$ be an arbitrary set of samples with the same dimension as $\boldsymbol{X}$. We define the \textbf{residual} (abbrev. \texttt{RES}) of $\boldsymbol{X}$ w.r.t. $\boldsymbol{Y}$ as a subset of $n$ samples of $\boldsymbol{X}$ such that, when concatenated with $\boldsymbol{Y}$, the empirical distribution of the concatenated samples is most similar to the data distribution $\mathbb{P}(\boldsymbol{x})$:
\begin{equation}
  \label{eq:residual}
  \texttt{RES}(\boldsymbol{X}, \boldsymbol{Y}, n) := \mathop{\arg\min}_{\boldsymbol{X}' \subseteq \boldsymbol{X}, |\boldsymbol{X}'| = n} d(\boldsymbol{X}, \boldsymbol{Y} \cup \boldsymbol{X}')
\end{equation}
where $d$ can be any distance metric between two empirical distributions.
\end{definition}

\begin{remark}
  In our case, $\boldsymbol{X}$ is the real tabular samples, and $\boldsymbol{Y}$ is the current generated samples by a LLM. Intuitively, the residual samples capture the part of the real samples that LLM has not yet grasped, thus named as \textit{residual}. To prevent overly long context prompts when interacting with the LLM, we enforce an upper-bound 
  $n$ on the size of the residual samples. 
  In our experiments, we set $n=500$ and instantiate $d$ as Jensen-Shannon Divergence (JSD) and Kolmogorov-Smirnov Distance (KSD).
\end{remark}

Since brute-force way of computing the residual is computationally prohibitive for large $N$ and $n$, we introduce a heuristic for sampling the residual. We describe details in the following—pseudo-code is provided in Appendix \ref{appendix:res_alg}.

\subsection{Compute Residual}
We propose to use a simple heuristic to shrink the search space. Specifically, we first randomly select a column, then we group the real samples $\boldsymbol{X}$ based on the value of the selected column\footnote{For categorical columns, we group by the categorical values. For continuous columns, we discretize them into a fixed number of bins and group by the bin index.}. Each group of samples is then concatenated with the generated samples $\boldsymbol{Y}$. Finally, we select the group that has the smallest distance to the real samples $\boldsymbol{X}$ as the residual. The time complexity of this heuristic search algorithm is $O(N)$. Additionally, the final residual subset always exhibits a consistent pattern—either sharing the same category or falling within a narrow numerical range in one of its columns. We hypothesize that this simple pattern makes the residual samples particularly effective as in-context examples for LLMs (see Figure \ref{fig:in-context-examples} (c)). 

\input{Tables/fidelity}


\subsection{Table Generation by \modelname}
\modelname can be easily integrated with LLMs to generate high-quality synthetic tabular data. See Fig. \ref{fig:overview} for an overview of the procedure. Here are the concrete steps involved in this procedure:
\begin{enumerate}[leftmargin=*]
  \item \textbf{In-context Prompting:}  For the first iteration, we randomly select $n$ samples from the real dataset $\boldsymbol{X}$ as the initial set of in-context examples.
  Otherwise, we plug the residual samples computed in the previous iteration into the prompt template to prompt LLMs. 
  We append the generated samples into $\boldsymbol{Y}$.
  \item \textbf{Residual Computation:} We then compute the residual of $\boldsymbol{X}$ w.r.t. $\boldsymbol{Y}$: $\texttt{RES}(\boldsymbol{X}, \boldsymbol{Y}, n)$. Specifically, if the current iteration is an even number, we instantiate $d$ as JSD, otherwise, we instantiate $d$ as KSD.
  \item \textbf{Iterative Refinement:} Repeat the above steps until enough synthetic samples are generated.
\end{enumerate}

%% file: Tables/fidelity.tex
\begin{table*}[!t]
\footnotesize
\setlength{\tabcolsep}{4.5pt}
\begin{center}
\begin{tabular}{llcccccccccc}
\toprule

& \textbf{Method} & \textbf{Marginal}$\downarrow \%$  & \textbf{Corr}$\downarrow \%$ & \textbf{Precision}$\downarrow \%$ & \textbf{Recall}$\downarrow \%$  & \textbf{C2ST}$\downarrow \%$ & \textbf{JSD}$\downarrow 10^{-2}$   \\
\midrule
\rowcolor{gray!20}
\multicolumn{8}{c}{\texttt{\textbf{VAE-based}}} \\
& TVAE \citep{ctgan} & $14.61$ & $17.32$ & $11.65$ & $9.11$ & $41.72$ & $0.63$ \\
\midrule 
\rowcolor{gray!20}
\multicolumn{8}{c}{\texttt{\textbf{GAN-based}}} \\
& CTGAN \citep{ctgan} & $16.36$ & $20.33$ & $30.65$ & $11.41$ & $42.90$ & $0.91$ \\
\midrule 
\rowcolor{gray!20}
\multicolumn{8}{c}{\texttt{\textbf{Diffusion-based}}} \\
& STaSy \citep{stasy} & $12.35$ & $9.72$ & $11.09$ & $2.66$ & $55.82$ & $1.34$ \\
& CoDi \citep{codi} & $21.70$ & $24.92$ & $9.89$ & $6.74$ & $57.88$ & $1.07$ \\
& TabDDPM \citep{tabddpm} & $14.04$ & $8.16$ & $13.37$ & $2.27$ & $24.21$ & $0.85$ \\
& TabSyn \citep{tabsyn} & $1.40$ & $2.36$ & $3.76$ & $2.29$ & $2.64$ & $0.05$ \\
\midrule
\rowcolor{gray!20}
\multicolumn{8}{c}{\texttt{\textbf{Autoregressive Models}}} \\
& RTF \citep{realtabformer} & $5.31$ & $10.42$ & $3.53$ & $5.25$ & $28.16$ & $0.45$ \\
& TabMT \citep{tabmt} & $4.46$ & $8.24$ & $33.88$ & $50.44$ & $44.77$ & $0.63$ \\
\midrule 
\rowcolor{gray!20}
\multicolumn{8}{c}{\texttt{\textbf{LLM-Finetuned}}} \\
& GReaT \citep{great} & $15.53$ & $40.48$ & $1.49$ & $10.06$ & $48.28$ & $1.06$ \\
\midrule 
\rowcolor{gray!20}
\multicolumn{8}{c}{\texttt{\textbf{LLM-Prompt-Only}}} \\
& CLLM w. GPT-4o-mini & $13.17$ & $19.57$ & $6.63$ &  $8.08$ & $39.02$ & $0.78$ \\

& \modelname w. GPT-4o-mini (\textbf{Ours}) & $11.39$ & $17.07$ & $5.54$ &  $4.67$ & $37.63$ & $0.80$ \\

& Improvement   & {\fontsize{10.5}{12.6}\selectfont $\redbf{13.5\%}$} & {\fontsize{10.5}{12.6}\selectfont $\redbf{12.8\%}$}  & {\fontsize{10.5}{12.6}\selectfont $\redbf{19.7\%}$}  & {\fontsize{10.5}{12.6}\selectfont $\redbf{42.2\%}$} & {\fontsize{10.5}{12.6}\selectfont $\redbf{3.5\%}$} & $-$\\

\midrule
& CLLM w. GPT-4o & $10.57$ & $13.46$ & $4.00$ &  $4.25$ & $31.51$ & $0.63$ \\
& \modelname w. GPT-4o (\textbf{Ours}) & $9.14$ & $12.86$ & $4.93$ &  $2.80$ & $26.70$ & $0.62$ \\
& Improvement   & {\fontsize{10.5}{12.6}\selectfont $\redbf{13.6\%}$} & {\fontsize{10.5}{12.6}\selectfont $\redbf{4.5\%}$}  & $-$  & {\fontsize{10.5}{12.6}\selectfont $\redbf{34.1\%}$} & {\fontsize{10.5}{12.6}\selectfont $\redbf{15.3\%}$} & {\fontsize{10.5}{12.6}\selectfont $\redbf{1.6\%}$}\\

\bottomrule
\end{tabular} 
\caption{\textbf{Fidelity comparison}: Comparison of various methods on fidelity metrics. Results are averaged over all datasets. All metrics are scaled to percentages ($\%$) or $10^{-2}$, and reversed so that the lower the better.}
\label{tbl:fidelity}
\end{center}
\vspace{-0.6em}
\vspace{-0.1in}
\end{table*}

%% file: Contents/4_experiment.tex
\section{Experiments}
\label{experiments}
\begin{figure*}[h!]
  \centering
  \vspace{-10pt}
  \includegraphics[width=1.0\linewidth,height=0.30\textheight]{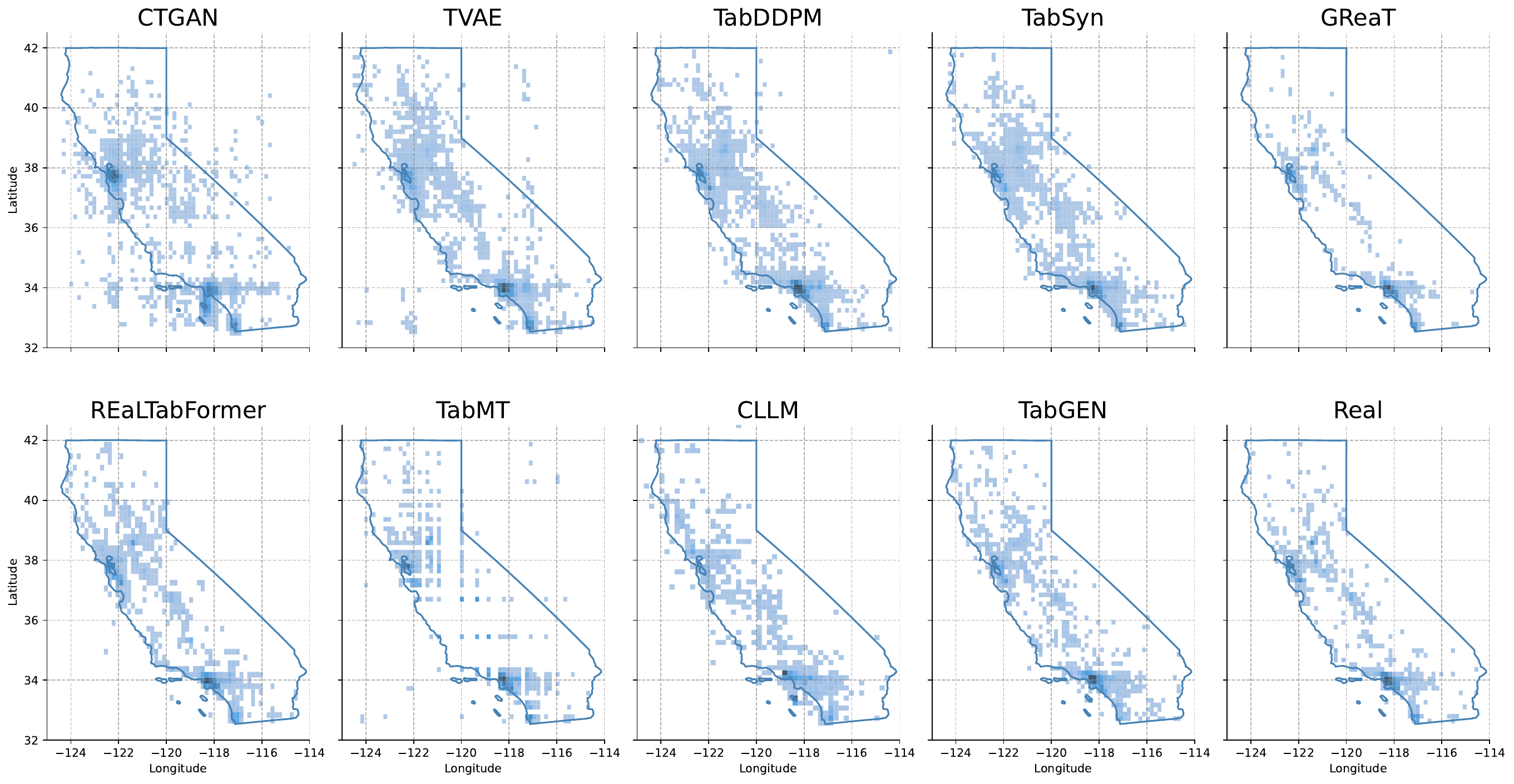}
  \caption{\textbf{Visual comparison}: 2D scatter plot of \texttt{Longitude} and \texttt{Latitude} attributes of California dataset. Real represents the original training datasets. All sets are downsampled to 3000 rows for better visualization.  \modelname generates spatially coherent synthetic data that closely matches the distribution of the original dataset.}
  \label{fig:2d-california}
  \vspace{-10pt}
\end{figure*}

We validate the performance of \modelname through extensive experiments. In particular, we investigate the following questions:
\begin{itemize}[leftmargin=*]
    \item Can \modelname improve generation quality compared to previous LLM-based methods? (Table. \ref{tbl:fidelity}, \ref{tbl:utility}, Fig. \ref{fig:2d-california})
    \item How does \modelname perform, compared to training-based deep generative models, under a data-scarcity setting? (Fig. \ref{fig:low-resource})
    \item Does \modelname generate new synthetic data instead of copying the training dataset? (Fig. \ref{fig:dcr})
\end{itemize}

\subsection{Setup}
\paragraph{Datasets.}
We select five real-world tabular datasets containing both numerical and categorical attributes: \textbf{Adult, Default, Shoppers, Magic} and \textbf{California}. The statistics of the datasets are summarized in Table \ref{tbl:stat-dataset} in Appendix \ref{appendix:datasets}.

\paragraph{Baselines.}
To comprehensively assess \modelname's performance, we conduct comparisons against a wide range of traditional deep generative models and LLM-based methods, which we categorize into the following two groups:
\begin{itemize}[leftmargin=*]
    \item \textbf{Deep generative models:} 1) VAE-based method TVAE \citep{ctgan}, 2) GAN-based method CTGAN \citep{ctgan}, 3) Diffusion-based method TabSyn \citep{tabsyn}, TabDDPM \citep{tabddpm}, CoDi \citep{codi}, STaSy \citep{stasy}, 4) Autoregressive method TabMT \citep{tabmt}, RealTabformer (RTF) \citep{realtabformer}.
    \item \textbf{LLM-based methods:} 1) with fine-tuning: GReaT \citep{great} 2) without fine-tuning: CLLM \cite{cllm}. CLLM was originally employed with GPT-3.5 and GPT-4, to ensure a fair comparison to CLLM, we employ CLLM with stronger models: GPT-4o-mini and GPT-4o, and we keep all the other experimental settings the same as ours.
\end{itemize}
To the best of our knowledge, CLLM \cite{cllm} is the only previous work that is training-free and solely based on in-context learning (excluding the curation step). \modelname falls in this setting.

\paragraph{Implementation details.}
Our main experiments employ GPT-4o-mini and GPT-4o as the LLMs. For all LLMs, we set the temperature to $1.0$. We generate 3000 samples ($N=3000$) for each dataset.
Each experiment is conducted 5 times and the average results are reported. 


\paragraph{Evaluation metrics.}
We evaluate the synthetic tabular data from three distinct dimensions: \circlednumber{1} \textit{Fidelity} - if the synthetic data faithfully recovers the ground-truth data distribution. We evaluate fidelity by 5 metrics: 1) Marginal distribution through Kolmogorov-Sirnov Test, 2) Pair-wise column correlation (Corr.) by computing Pearson Correlation, 3) Classifier Two Sample Test (C2ST) 4) Precision and Recall,  5) Jensen-Shannon Divergence (JSD). \circlednumber{2} \textit{Utility} - the utility of the synthetic data when used to train downstream models, we use the Train-on-Synthetic-then-Test (TSTR) protocol to evaluate the AUC score of XGBoost model on predicting the target column of each dataset. \circlednumber{3} \textit{Privacy} - if the synthetic data is not copied from the real records, we employ the Distance to Closest Record (DCR) metric.
We defer the full description of the metrics to Appendix \ref{appendix:metrics}.

Notably, previous works \citep{great, cllm} on evaluating LLMs for tabular data generation focus only on Machine Learning Utility and Privacy protection. Our paper fills this gap by providing the first comprehensive evaluation of LLMs' ability on tabular data synthesis.

\input{Tables/utility}

\subsection{\modelname outperforms LLM-based baseline methods} \label{sec:bsl}

As shown in Table \ref{tbl:fidelity}, \modelname consistently outperforms current LLM-based approaches on fidelity metrics, including both the training-free method CLLM and the fine-tuning-based method GReaT. Specifically, when using GPT-4o-mini, \modelname improves fidelity scores by a margin of 3.5$\%–42.2\%$, and when using GPT-4o, the improvements range from $1.6\%$ to 34.1\% across various metrics. Notably, the highest gains are observed in Recall: 42.2\% improvement with GPT-4o-mini and 34.1\% with GPT-4o.
Recall measures whether the synthetic data adequately covers the diverse spectrum of the real data; thus, improved Recall signifies enhanced diversity in the synthesized samples. This significant improvement is attributable to \modelname’s strategy of computing residual samples at each prompt iteration. These residual samples target underrepresented regions of the data distribution, thereby enriching the overall diversity of the synthetic data. This observation further validates the effectiveness of \modelname’s residual-based iterative refinement mechanism.

\subsection{\modelname outperforms deep generative models under data-scarcity}

\begin{figure}[h]
  \centering
  \includegraphics[width=0.95\linewidth]{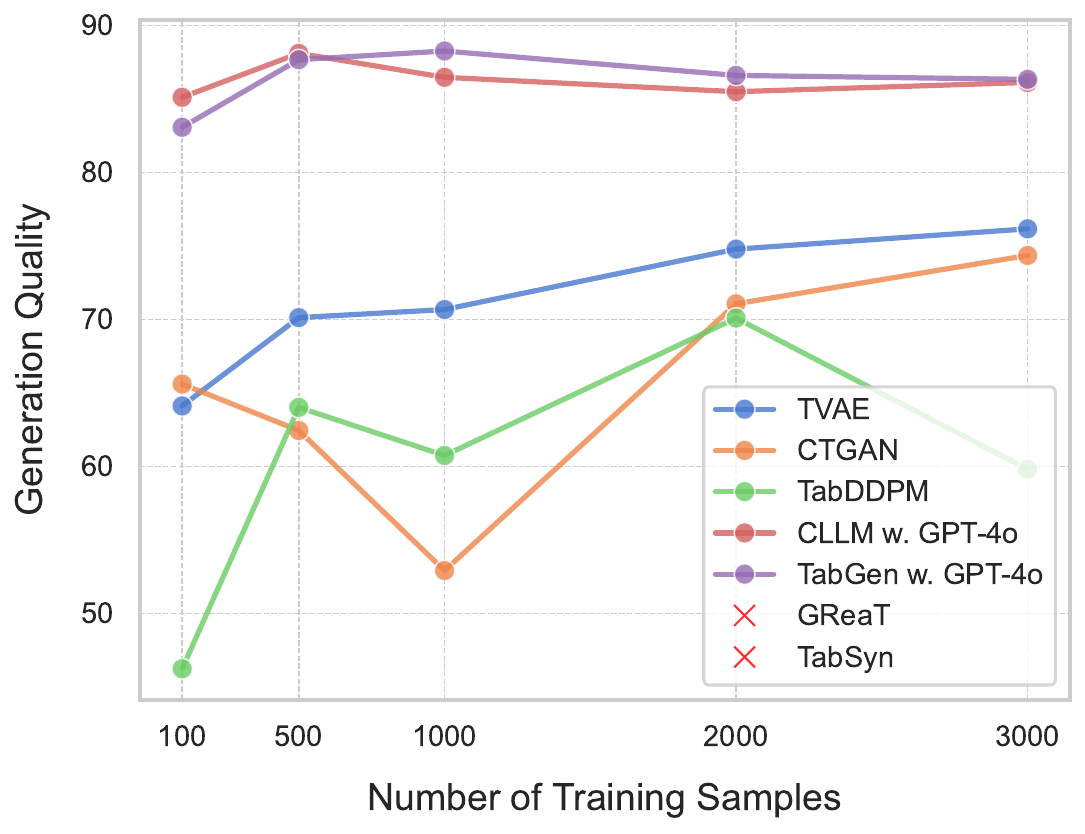}
  \caption{\textbf{Quality comparison under data-scarcity}. \modelname and CLLM achieves the highest quality score under the few-shot setting. TabSyn and GReaT fail to generate meaningful data.}
  \label{fig:low-resource}
\end{figure}

One important application of tabular data synthesis is addressing data scarcity. In many cases, we have access to only a limited number of real data points, yet we require a much larger dataset to adequately train our downstream models. To generate sufficient training data, generative models can be employed.
In our experiments, we evaluate the performance of \modelname in comparison with other deep generative models under data-scarce conditions. To simulate such scenarios, we created training sets by randomly sampling 100, 500, 1000, 2000, and 3000 rows from the Default dataset. The generative models were then trained on these subsets, and the synthesized data’s quality was evaluated using the original full training set of 30,000 rows.
As shown in Figure \ref{fig:low-resource}, deep generative models like TVAE, CTGAN, and TabDDPM exhibit a significant drop in performance when trained on limited data. In contrast, \modelname and CLLM maintain performance comparable to the full-data setting. This is attributed to the strong prior distribution provided by large language models (LLMs). Notably, the performance between \modelname and CLLM is very similar because, in data-scarce scenarios, the entire training set can be used as in-context learning examples for the LLMs. Hence, the \textit{residual} sampling effectively degenerates to random sampling.

\subsection{\modelname does not copy training data}
\begin{figure*}[t!]
  \centering
  \includegraphics[width=1.0\textwidth,angle=0]{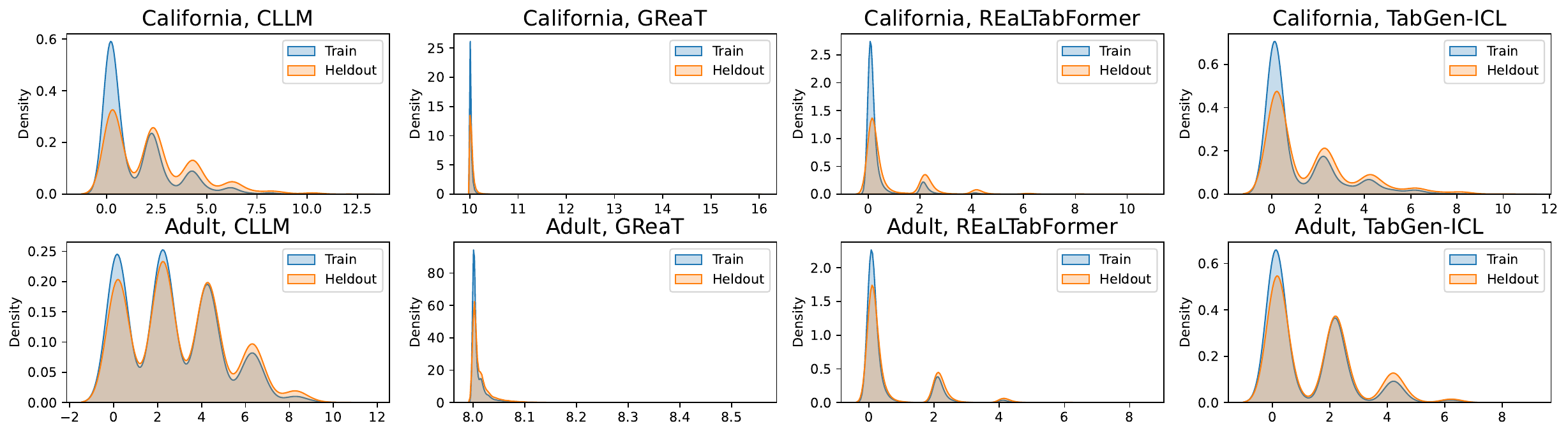}
  \caption{\textbf{Privacy comparison}: Distributions of the DCR scores between the synthetic dataset and the training/holdout datasets. \modelname and Curated-LLM (CLLM) are both employed with GPT-4o-mini. \label{fig:dcr}}
\end{figure*}

In Figure \ref{fig:dcr}, we illustrate the distributions of the L2 distances between the synthetic data and both the training and holdout datasets for CLLM, GReaT, REaLTabFormer (RTF), and \modelname. Notably, \modelname exhibits nearly identical distributions for the training and holdout sets, suggesting it is less prone to copying the training data. In contrast, CLLM and GReaT show more disparate distributions, indicating a higher likelihood of relying on the training data.

\vspace{-2em}

\vspace{16pt}

\subsection{Ablation Study}
\paragraph{Effect of $d$.} We examine the effect of the distribution distance metric $d$ used for quantifying residual in Equation \ref{eq:residual}. We test \modelname (w. GPT-4o-mini) with only KSD or JSD metric and compare it with our alternating strategy (KSD+JSD) on the California dataset. As shown in Table \ref{tbl:distance}, the alternating strategy achieves the best performance. 

\begin{table}[h]
\vspace{-2mm}
\centering
\begin{tabularx}{\linewidth}{%
  >{\centering\arraybackslash}X | 
  >{\centering\arraybackslash}X | 
  >{\centering\arraybackslash}X | 
  >{\centering\arraybackslash}X%
}
\toprule
\textbf{$d$} & KSD & JSD & KSD+JSD \\
\midrule
Marg. & $92.43${\tiny$\pm 0.005$} & 90.72{\tiny$\pm 0.009$} & $92.48${\tiny$\pm 0.008$} \\[1mm]
Corr. & $88.41${\tiny$\pm 0.022$} & 90.67{\tiny$\pm 0.021$} & $91.24${\tiny$\pm 0.016$} \\
\bottomrule
\end{tabularx}
\caption{Ablation study for $d$.}
\label{tbl:distance}
\end{table}
\vspace{-1.0em}

\paragraph{Effect of Large Language Models.}
In this section, we investigate the impact of large language model (LLM) capabilities on \modelname's performance. We evaluate \modelname using LLMs of varying parameter sizes, including Gemini-1.5-Flash, Gemini-1.5-Pro \citep{gemini}, Claude-3-Haiku, Claude-3-Sonnet \citep{claude3}, LLaMA-3.1 8B, LLaMA-3.1 405B \citep{llama3}, and Qwen2 \citep{qwen2}. We assess the marginal metric on the California dataset, with results presented in Table \ref{tbl:ablation_model}. Our findings reveal a correlation between LLM capacity and synthetic data generation quality. As the LLMs' capacity increases, the quality of generated synthetic data improves. We hypothesize that this improvement stems from larger models' enhanced ability to capture and reproduce complex patterns within the data, resulting in more realistic synthetic outputs. This relationship underscores the importance of model capacity in generating high-quality synthetic data.

\input{Tables/ablation_model}



%% file: Tables/utility.tex
\begin{table*}[!t]
\footnotesize
\setlength{\tabcolsep}{4.5pt}
\begin{center}
\begin{tabular}{llcccccccccc}
\toprule

& \multirow{2}{*}{\textbf{Method}} & \textbf{California} & \textbf{Adult} & $\textbf{Shoppers}$ & $\textbf{Magic}$  & \textbf{Default} &  \\
& & AUC$\uparrow$ & AUC$\uparrow$ & AUC$\uparrow$ & AUC$\uparrow$ & AUC$\uparrow$ \\
\midrule
& Real & $0.999$ & $0.927$ & $0.926$ & $0.946$ & $0.770$  \\
\midrule
\rowcolor{gray!20}
\multicolumn{8}{c}{\texttt{\textbf{VAE-based}}} \\
& TVAE \citep{ctgan} & $0.986$ & $0.846$ & $0.898$ & $0.912$ & $0.744$  \\
\midrule 
\rowcolor{gray!20}
\multicolumn{8}{c}{\texttt{\textbf{GAN-based}}} \\
& CTGAN \citep{ctgan} & $0.925$ & $0.874$ & $0.868$ & $0.874$ & $0.736$  \\
\midrule 
\rowcolor{gray!20}
\multicolumn{8}{c}{\texttt{\textbf{Diffusion-based}}} \\
& STaSy \citep{stasy} & $0.997$ & $0.903$ & $0.909$ & $0.923$ & $0.749$ \\
& CoDi \citep{codi} & $0.981$ & $0.829$ & $0.855$ & $0.930$ & $0.497$ \\
& TabDDPM \citep{tabddpm} & $0.992$ & $0.911$ & $0.915$ & $0.933$ & $0.763$ \\
& TabSyn \citep{tabsyn} & $0.993$ & $0.904$ & $0.913$ & $0.934$ & $0.764$ \\

\midrule 
\rowcolor{gray!20}
\multicolumn{8}{c}{\texttt{\textbf{Autoregressive Models}}} \\
& RTF \citep{realtabformer} & $0.948$ & $0.925$ & $-$ & $0.931$ & $0.764$ \\
& TabMT \citep{tabmt} & $0.988$ & $0.873$ & $0.912$ & $0.822$ & $0.714$ \\

\midrule 
\rowcolor{gray!20}
\multicolumn{8}{c}{\texttt{\textbf{LLM-Finetuned}}} \\
& GReaT \citep{great} & $0.996$ & $0.913$ & $0.902$ & $0.888$ & $0.755$ \\

\midrule 
\rowcolor{gray!20}
\multicolumn{8}{c}{\texttt{\textbf{LLM-Prompt-Only}}} \\
& CLLM w. GPT-4o-mini & $0.840$ & $0.879$ & $0.708$ & $0.826$ & $0.557$ \\

& \modelname w. GPT-4o-mini (\textbf{Ours}) & $0.947$ & $0.894$ & $0.792$ & $0.891$ & $0.628$  \\
& Improvement   & {\fontsize{10.5}{12.6}\selectfont $\redbf{12.7\%}$} & {\fontsize{10.5}{12.6}\selectfont $\redbf{1.7\%}$}  & {\fontsize{10.5}{12.6}\selectfont $\redbf{11.9\%}$}  & {\fontsize{10.5}{12.6}\selectfont $\redbf{7.9\%}$} & {\fontsize{10.5}{12.6}\selectfont $\redbf{12.7\%}$} \\

\midrule 
& CLLM w. GPT-4o & $0.947$ & $0.891$ & $0.865$ & $0.885$ & $0.718$ \\

& \modelname w. GPT-4o (\textbf{Ours}) & $0.975$ & $0.892$ & $0.879$ & $0.903$ & $0.713$ \\

& Improvement   & {\fontsize{10.5}{12.6}\selectfont $\redbf{3.0\%}$} & {\fontsize{10.5}{12.6}\selectfont $\redbf{0.1\%}$}  & {\fontsize{10.5}{12.6}\selectfont $\redbf{1.4\%}$}  & {\fontsize{10.5}{12.6}\selectfont $\redbf{1.8\%}$} & {\fontsize{10.5}{12.6}\selectfont $\redbf{0.5\%}$} \\
\bottomrule
\end{tabular} 
\caption{\textbf{Utility comparison}: AUC scores of Train-on-synthetic-Test-on-real (TSTR) XGBoost model predicting the target column of each table. $\uparrow$ indicates the higher the better. $-$ indicates training failure.}
\label{tbl:utility}
\end{center}
\vspace{-0.6em}
\vspace{-0.1in}
\end{table*}

%% file: Tables/ablation_model.tex
\begin{table}[h]
    \vspace{-1mm}
    \centering
    \begin{tabularx}{\linewidth}{>{\centering\arraybackslash}X | >{\centering\arraybackslash}X | c}
        \toprule
        \textbf{Model} & \textbf{Quality$\uparrow (\%)$} & \textbf{Rank} \\
        \midrule
        GPT-4o-mini       & $91.86${\tiny$\pm 0.008$}   & 3 \\
        GPT-4o            & $94.69${\tiny$\pm 0.006$}  & 1 \\
        Gemini-1.5-Flash  & $89.96${\tiny$\pm 0.017$}  & 4 \\
        Gemini-1.5-Pro    & $92.21${\tiny$\pm 0.033$}  & 2 \\
        Claude-3-Haiku    & $89.17${\tiny$\pm 0.031$}  & 5 \\
        Claude-3-Sonnet   & $88.97${\tiny$\pm 0.068$}  & 6 \\
        \bottomrule
    \end{tabularx}
    \caption{Effect of large language models on the performance of \modelname.}
    \vspace{-1em}
    \label{tbl:ablation_model}
\end{table}

%% file: Contents/5_conclusion.tex
\section{Conclusion}
This work proposes \modelname, an ICL framework for tabular data generation with LLMs. It steer the LLM's prior distribution towards real data distribution by iteratively retrieving the most under-represented regions. Extensive experiments validate the effectiveness of our approach, demonstrating its potential to enhance LLM-based synthetic data generation across various domains.

%% file: Contents/6_limitations.tex
\section{Limitations}

\modelname relies on a simple heuristic search algorithm to compute \textit{residual} samples, under which the optimality is not guaranteed. In future, we will explore more principled approaches to compute residuals.

%% file: Contents/_appendix.tex
\onecolumn
\section{Appendix} \label{sec:appendix}

\subsection{Prompts used for generating tabular data} \label{lst:prompt}
This prompt template is used in Section~\ref{method} to generate realistic data that follows the same distribution as the given real data.

\begin{lstlisting} 
You are a synthetic data generator tasked with creating new tabular data samples that closely mirror the distribution and characteristics of the original dataset.

# Instruction
1. Analyze the provided real samples carefully.
2. Generate synthetic data that maintains the statistical properties of the real data.
3. Ensure all attributes cover their full expected ranges, including less common or extreme values.
4. Maintain the relationships and correlations between different attributes.
5. Preserve the overall distribution of the real data while introducing realistic variations.

# Key points to consider
- Replicate the data types of each column (e.g., numerical, categorical).
- Match the range and distribution of numerical attributes.
- Maintain the frequency distribution of categorical attributes.
- Reflect any patterns or trends present in the original data.
- Introduce realistic variability to avoid exact duplication.

# Real samples
{data}

# Output format:
Please present the generated data in a JSON format, structured as a list of objects, where each object represents a single data point with all attributes.

\end{lstlisting}

\subsection{Dummy Prompt} \label{appendix:dummy_prompt}
The following prompt only contains the column names, but not any actual data in it. It is used to produce the results in Fig.\ref{fig:in-context-examples} (a).
\begin{lstlisting}
You are a synthetic data generator tasked with creating new tabular data samples that closely mirror the distribution and characteristics of the original dataset.
Generate 50 samples of synthetic data.
    
Each sample should include the following attributes:
{attributes_list}

Make sure that the numbers make sense for each attribute. 

Output Format:
Present the generated data in a JSON format, structured as a list of objects, where each object represents a single data point with all attributes.

\end{lstlisting}

\subsection{JSON Schema} \label{lst:json}
The following code define the JSON data class for the structured output function of GPT-4o and GPT-4o-mini.
\begin{lstlisting}
def create_json_model(df: pd.DataFrame, dataname=None) -> BaseModel:
    fields = {}
    
    for column in df.columns:
        if df[column].dtype == 'object':  
            fields[column] = (str, ...)
        elif df[column].dtype == 'int64':
            fields[column] = (int, ...)
        elif df[column].dtype == 'float64':
            fields[column] = (float, ...)
        elif df[column].dtype == 'bool':
            fields[column] = (bool, ...)
        else:
            raise TypeError(f"Unexpected dtype for column {column}: {df[column].dtype}")
    
    JSONModel = create_model(dataname, **fields)
    
    class JSONListModel(BaseModel):
        JSON: List[JSONModel]

    return JSONListModel
\end{lstlisting}

\subsection{Heuristic for computing residual}
In this section, we provide the pseudo-code of our heuristic strategy for computing the residual.
\begin{algorithm}
\caption{Compute residual \label{appendix:res_alg}}
\begin{algorithmic}[1]
\Require current dataset $X$, target dataset $Y$, distribution distance $d$.
\State Randomly select a column index $j$
\If{column $j$ is categorical}
    \State Let $C_j$ be the number of categories in column $j$
    \State Group samples in $Y$ into $C_j$ number of subsets based on its category on column $j$, denote the set of subsets by $(Y_j^i)_{i=1}^{C_j}$
\Else
    \State Quantize column $i$ into 50 bins
    \State $C_j\leftarrow$ 50
    \State Group samples in $Y$ into $C_j$ number of subsets based on its bin index on column $j$, denote the set of subsets by $(Y_j^i)_{i=1}^{C_j}$
\EndIf
\For{$i = 1$ to $C_j$} 
    \State Compute distance between $Y_j^i \cup X$ and $Y$: $d_i = d(Y_j^i \cup X, Y)$
\EndFor
\State \Return subset $Y_j^i$ that attains the minimal distance.
\end{algorithmic}
\end{algorithm}

\subsection{Datasets} \label{appendix:datasets}
We use five real-world datasets of varying scales, and all of them are available at Kaggle\footnote{\url{https://www.kaggle.com/}} or the UCI Machine Learning repository\footnote{\url{https://archive.ics.uci.edu/}}. We consider five datasets containing both numerical and catergorical attributes: California\footnote{\url{https://www.kaggle.com/datasets/camnugent/california-housing-prices}}, Magic\footnote{\url{https://archive.ics.uci.edu/dataset/159/magic+gamma+telescope}}, Adult\footnote{\url{https://archive.ics.uci.edu/dataset/2/adult}}, Default\footnote{\url{https://archive.ics.uci.edu/dataset/350/default+of+credit+card+clients}}, Shoppers\footnote{\url{https://archive.ics.uci.edu/dataset/468/online+shoppers+purchasing+intention+dataset}}. The statistics of these datasets are presented in Table~\ref{tbl:stat-dataset}. 
\begin{table}[h] 
    \centering
    \caption{Statistics of datasets. \# Num stands for the number of numerical columns, and \# Cat stands for the number of categorical columns.} 
    \label{tbl:stat-dataset}
    \small
    \begin{threeparttable}
    {
    \scalebox{0.95}
    {
	\begin{tabular}{lccccccccc}
            \toprule[0.8pt]
            \textbf{Dataset}  &  \# Rows  & \# Num & \# Cat & \# Train  & \# Test  \\
            \midrule 
            \textbf{California} Housing  & $20,640$ & $9$ & 1 & $18,390$ & $2,250$   \\
            \textbf{Magic} Gamma Telescope & $19,021$ & $10$ & 1 & $17,118$ & $1,903$  \\
            \textbf{Adult} Income & $32,561$ & $6$ & $8$ & $22,792$ & $9,769$ \\
            \textbf{Default} of Credit Card Clients & $30,000$ & $14$ & $10$ & $27,000$ & $3,000$\\
            Online \textbf{Shoppers} Purchase & $12,330$ & $10$ & $7$ & $11,098$ & $1,232$ \\
		\bottomrule[1.0pt] 
		\end{tabular}
   }
  }        
  \end{threeparttable}
\end{table}

\subsection{Evaluation Metrics} \label{appendix:metrics}
\paragraph{Fidelity}
To evaluate if the generated data can faithfully recover the ground-truth data distribution, we employ the following metrics: 
1) \textbf{Marginal distribution:} The Marginal metric evaluates if each column's marginal distribution is faithfully recovered by the
synthetic data. We use Kolmogorov-Sirnov Test for continuous data and Total Variation Distance for discrete data.
2) \textbf{Pair-wise column correlation}: This metric evaluates if the correlation between every two columns in the real data is captured by the synthetic data. We compute the Pearson Correlation between all pairs of columns then take average. In addition, we present joint density plots for the Longitude and Latitude features in the California Housing data set in Figure \ref{fig:2d-california}.  
3) \textbf{Classifier Two Sample Test (C2ST):} This metric evaluates how difficult it is to distinguish real data from synthetic data. Specifically, we create an augmented table that has all the rows of real data and all the rows of synthetic data. Add an extra column to keep track of whether each original row is real or synthetic. Then we train a Logistic Regression classifier to distinguish real and synthetic rows. 
4) \textbf{Precision and Recall:} Precision measures the quality of generated samples. High precision means the generated samples are realistic and similar to the true data distribution. Recall measures how much of the true data distribution is covered by the generated distribution. High recall means the model captures most modes/variations present in the true data.
5) \textbf{Jensen-Shannon Divergence (JSD):} This metric evaluates the Jensen-Shannon divergence \citep{jsd} between the distributions of real data and synthetic data.

\paragraph{Utility} 
We evaluate the utility of the generated data by assessing their performance in Machine Learning Efficiency (MLE). Following the previous works \cite{tabsyn},  we first split a real table into a real training and a real testing set. The generative models are trained on the real training set, from which a synthetic set of equivalent size is sampled. This synthetic data is then used to train a classification/regression model (XGBoost Classifier and XGBoost Regressor~\citep{xgboost}), which will be evaluated using the real testing set. The performance of MLE is measured by the AUC score for classification tasks and RMSE for regression tasks.

\paragraph{Privacy}
A high-quality synthetic dataset should accurately reflect the underlying distribution of the original data, rather than merely replicating it. To assess this, we employ the Distance to Closest Record (DCR) metric. We begin by splitting the real data into two equal parts: a training set and a holdout set. Using the training set, we generate a synthetic dataset. We then measure the distances between each synthetic data point and its nearest neighbor in both the training and holdout sets. In theory, if both sets are drawn from the same distribution, and if the synthetic data effectively captures this distribution, we should observe an equal proportion (around 50$\%$) of synthetic samples closer to each set. However, if the synthetic data simply copies the training set, a significantly higher percentage would be closer to the training set, well exceeding the expected $50\%$.

\subsection{Scalability of \modelname}
To evaluate the scalability of \modelname, we compare \modelname with CLLM on a large-scale dataset: Covertype dataset. This dataset consists of 581,012 instances and 54 features. \modelname and CLLM use iterative in-context learning to generate samples, thus the running time of these two methods are agnostic to the size of the training dataset, making them scalable to large datasets. In the following table, we compare \modelname with CLLM, employed with both GPT-4o mini and GPT-4o. 
\begin{table}[!t]
\begin{center}
\caption{Performance Comparison on the \textbf{Covertype} Dataset. For all the metrics except AUC, we scale the metrics to a base of 
$10^{-2}$, and reverse it so that lower values indicate better performance. For AUC, the higher the better.}
\begin{tabular}{lrrrrrrr}
\hline
Model & Marginal $\downarrow$ & Corr $\downarrow$ & C2ST $\downarrow$ & Precision $\downarrow$ & Recall $\downarrow$ & JSD $\downarrow$ & AUC $\uparrow$ \\
\hline
\hline
CLLM w. GPT-4o & 3.28 & 10.70 & 55.67 & 32.32 & 1.95 & 0.6078 & 0.8822 \\
TabGEN w. GPT-4o & 2.83 & 11.10 & 49.91 & 24.03 & 1.27 & 0.5109 & 0.9070 \\
Improvement (\%) & $\redbf{13.72\%}$ & - & $\redbf{10.34\%}$ & $\redbf{25.62\%}$ & $\redbf{34.87\%}$ & $\redbf{15.94\%}$ & $\redbf{2.81\%}$ \\
\hline
\hline
CLLM w. GPT-4o mini & 5.35 & 13.70 & 0.7796 & 0.3225 & 0.0591 & 0.8743 & 0.7113 \\
TabGEN w. GPT-4o mini & 5.09 & 12.62 & 0.7554 & 0.2876 & 0.0436 & 0.9353 & 0.8311 \\
Improvement (\%) & $\redbf{4.86\%}$ & $\redbf{7.88\%}$ & $\redbf{3.11\%}$ & $\redbf{10.81\%}$ & $\redbf{26.25\%}$ & - & $\redbf{16.86\%}$ \\
\hline
\end{tabular}
\end{center}
\end{table}

The results demonstrate that \modelname outperforms CLLM on most of the metrics. Notably, the Recall metric again shows the greatest improvement: 34.85\% on GPT-4o and 26.25\% on GPT-4o mini. This observation is consistent with our original findings in Sec. \ref{sec:bsl}. We believe these results strongly support \modelname's scalability to larger, more complex datasets.